\documentclass{article} 
\usepackage{iclr2022_conference,times}

\usepackage{amsmath,amsfonts,bm}

\def\eqref#1{equation~\ref{#1}}

\def\1{\bm{1}}

\DeclareMathAlphabet{\mathsfit}{\encodingdefault}{\sfdefault}{m}{sl}
\SetMathAlphabet{\mathsfit}{bold}{\encodingdefault}{\sfdefault}{bx}{n}

\usepackage{hyperref}
\usepackage{url}
\usepackage{booktabs}       
\usepackage{amsfonts}       
\usepackage{nicefrac}       
\usepackage{microtype}      
\usepackage{xcolor}
\usepackage{subcaption}
\usepackage{amsmath,bm,multicol,multirow}
\usepackage{cleveref}
\usepackage{adjustbox}
\usepackage{pgfplotstable}
\usepackage{supertabular} 
\usepackage{hyphenat}
\usepackage{makecell}

\usepgfplotslibrary{fillbetween}
\usetikzlibrary[patterns]

\title{XLS-R: Self-supervised Cross-lingual Speech Representation Learning at Scale}

\author{%
\centerline{
Arun Babu$^{\bigtriangleup}$\thanks{Equal contribution.}~, 
Changhan Wang$^{\bigtriangleup}$\samethanks~,
Andros Tjandra$^{\bigtriangleup}$, 
Kushal Lakhotia$^{\Diamond}$\thanks{Work done while at Facebook AI.}~, 
Qiantong Xu$^{\bigtriangleup}$,} \vspace{1mm}\\
\centerline{\textbf{
Naman Goyal$^{\bigtriangleup}$,
Kritika Singh$^{\bigtriangleup}$, 
Patrick von Platen$^{\clubsuit}$, 
Yatharth Saraf$^{\bigtriangleup}$,
Juan Pino$^{\bigtriangleup}$,}}\vspace{1mm}\\
\centerline{\textbf{
Alexei Baevski$^{\bigtriangleup}$, 
Alexis Conneau$^{\Box}$\thanks{Equal advising.}~,
Michael Auli$^{\bigtriangleup}$\samethanks}}\vspace{1mm}\\
\centerline{\normalfont $^{\bigtriangleup}$~Meta AI\hspace{0.2in}
$^{\Box}$~Google AI\hspace{0.2in}
$^{\Diamond}$~Outreach\hspace{0.2in}
$^{\clubsuit}$~Hugging Face}
}

\newcommand{\insertBABELtable}{
\begin{table}[h]
\centering
\captionof{table}{Speech recognition results on BABEL in terms of word error rate (WER) on  Assamese (as), Tagalog (tl), Swahili (sw), Lao (lo) and Georgian (ka). 
\label{tab:bbl_sota}}            
\begin{tabular}[t]{lrrrrr}
\toprule
& as & tl & sw & lo & ka \\
\midrule
\multicolumn{1}{l}{Labeled data} & 55h & 76h & 30h & 59h & 46h \\
\midrule
\multicolumn{1}{l}{\it Previous work} \\
\citet{alumae2016georgian} & - & - & - & - & 32.2 \\
\citet{ragni2018confidence} & - & 40.6 &  35.5 & - & - \\
\citet{inaguma2019transfer} & 49.1 & 46.3 & 38.3 & 45.7 & - \\
XLSR-10~\citep{conneau2021xlsr} & 44.9 & 37.3 &  35.5 &  32.2 & - \\
XLSR-53~\citep{conneau2021xlsr} & 44.1 & 33.2 & 26.5 & - &  31.1\\
\midrule
\multicolumn{1}{l}{\it This work} \\
\xlsrpb{0.3} & 42.9 & 33.2 & 24.3 &	31.7 & 28.0\\
\xlsrpb{1}   & 40.4 & 30.6 & 21.2 & 30.1 & 25.1 \\
\xlsrpb{2}   & \bf 39.0 & \bf 29.3 & \bf 21.0 & \bf 29.7 & \bf 24.3  \\
\bottomrule
\end{tabular}
\end{table}
}

\newcommand{\insertCVtable}{
\begin{table}[t]
\begin{center}
\caption{Phoneme recognition performance on CommonVoice in terms of phoneme error rate (PER) when using one hour of labeled data to fine-tune each language. 
We compare to m-CPC~\citep{rivire2020unsupervised}, \citet{fer2017multilingually}, XLSR-10~\citep{conneau2021xlsr} and XLSR-53~\citep{conneau2021xlsr}.
\label{tab:cv_results}}
\resizebox{1\linewidth}{!}{
    \begin{tabular}[b]{lcrrrrrrrrrrr}
        \toprule
        & es & fr & it & ky & nl & ru & sv & tr & tt & zh-HK & Avg \\
        \midrule
        Labeled data & 1h & 1h & 1h & 1h & 1h & 1h & 1h & 1h & 1h & 1h &   \\
        \midrule
        \multicolumn{9}{l}{\it Previous work} \\
        m-CPC & 38.0 & 47.1 & 40.5 & 41.2 & 42.5 & 43.7 & 47.5 & 47.3 & 42.0 & 55.0 & 44.5 \\
        \citet{fer2017multilingually} & 36.6 & 48.3 & 39.0 & 38.7 & 47.9 & 45.2 & 52.6 & 43.4 & 42.5 & 54.3 & 44.9 \\
        XLSR-10 & 7.9 & 12.6 & 11.7 & 7.0 & 14.0 & 9.3 & 20.6 & 9.7 & 7.2 & 22.8 & 12.3 \\
        XLSR-53 & 2.9 & 5.0 & 5.7 & 6.1 & 5.8 & 8.1 & 12.2 & 7.1 & 5.1 & 18.3 & 7.6 \\
        \midrule
        \multicolumn{2}{l}{\it This work} \\
        \xlsrpb{0.3} & 3.1 & 5.4&4.9&5.1 &	5.8&	6.0&	7.2&	6.0&	4.1 & 17.0 & 6.5\\
        \xlsrpb{1} &\bf 2.0 & \bf 3.9 & \bf 3.5 & 4.1 & \bf 4.2 & 4.1 & 5.5 & 4.4 & 3.4 & 15.7 & 5.1\\
        \xlsrpb{2} & 2.2 & 4.0& \bf 3.5 & \bf 4.0 & 4.7 & \bf 3.7 & \bf 5.0 & \bf 4.0 & \bf 2.9 & \bf 14.8 & \bf 4.9\\
        \bottomrule
    \end{tabular}
}
\end{center}
\end{table}
}

\newcommand{\insertVPtable}{
\begin{table}[h!]
\centering
\caption{VoxPopuli ASR results in terms of WER. We report the supervised-only baseline (No pretraining) of~\citet{wang2021vp} as well as their pretrained model (VP-10K).}
\begin{tabular}{l|rrrrrrrr}
\toprule
& en & de & it & fr & es & pl & ro & hu  \\
\midrule
Labeled data & 543h & 282h & 91h & 211h & 166h & 111h & 89h & 63h \\
\midrule 
\multicolumn{8}{l}{\it Baselines from previous work \citep{wang2021vp}} \\
No pretraining & 30.0 & 29.3 & 45.2 & 30.5 & 31.4 & 25.6 & 27.7 & 27.9 \\
VP-10K & 16.2 & 16.2 & 21.5 & 15.4 & 11.0 & 12.5 & 9.4 & 12.0 \\
\midrule
\multicolumn{8}{l}{\it This work} \\
\xlsrpb{0.3} & 10.2 & 13.0 & 19.2 & 12.6 & 9.8 & 9.6 & 7.9 & 11.6 \\
\xlsrpb{1}  & \bf  8.8 & \bf 11.5 & \bf 15.1 & \bf 10.8 & \bf 8.2 & \bf 7.7 & \bf 7.3 & \bf 9.6 \\
\bottomrule
\toprule
& nl & cs & sl & fi & hr & sk & Avg \\
\midrule
Labeled data & 53h & 62h & 10h & 27h & 43h & 35h & \\
\midrule 
\multicolumn{8}{l}{\it Baselines from previous work \citep{wang2021vp}} \\
No pretraining & 38.3 & 27.7 & 96.5 & 41.6 & 40.2 & 32.7 & 37.5 \\
VP-10K & 19.7 & 11.8 & 26.1 & 17.1 & 14.1 & 11.1 & 15.3 \\
\midrule
\multicolumn{8}{l}{\it This work} \\
\xlsrpb{0.3} & 14.8 & 10.5 & 24.5 & 14.2 & 12.3 & 8.9 & 12.8\\
\xlsrpb{1}  & \bf 12.5 & \bf 8.7 & \bf 19.5 & \bf 11.3 & \bf 10.0 & \bf 7.1 & \bf 10.6 \\
\bottomrule
\end{tabular}    
\label{tab:vp_asr}
\end{table}
}

\newcommand{\insertXLSR}{
\begin{figure*}[t]
	\begin{center}
          \includegraphics[width=\linewidth]{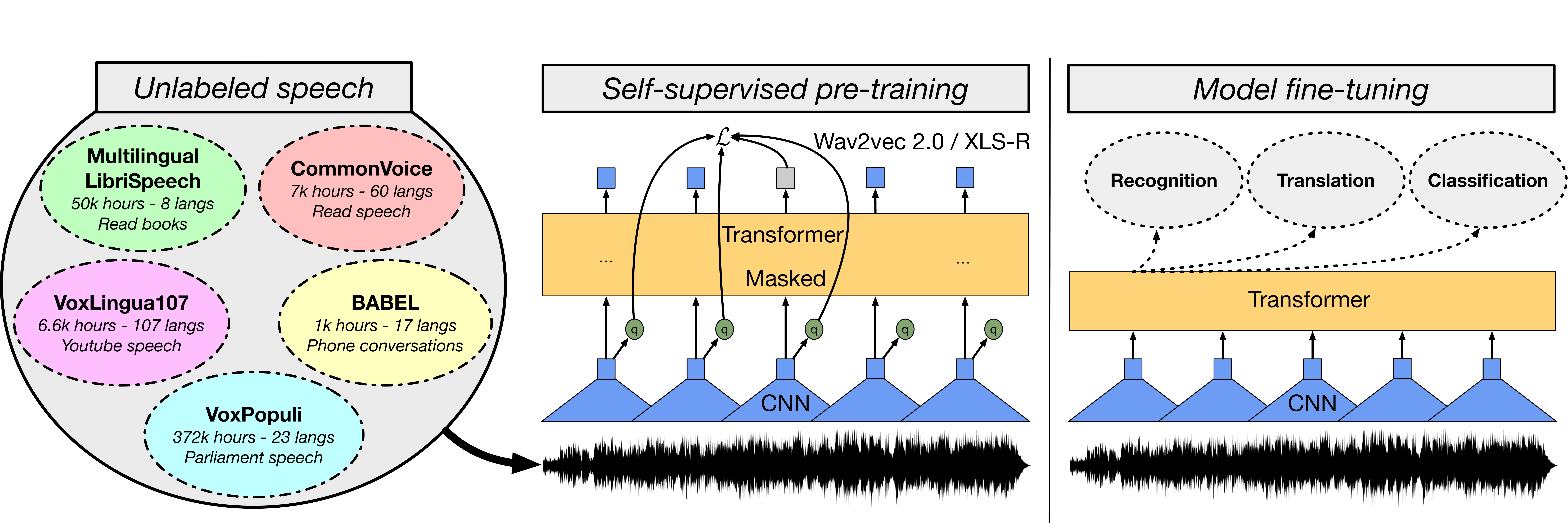}
        \captionof{figure}{\textbf{Self-supervised cross-lingual representation learning.} We pre-train a large multilingual wav2vec 2.0 Transformer (\xlsrp{}) on 436K hours of unannotated speech data in 128 languages. The training data is from different public speech corpora and we fine-tune the resulting model for several multilingual speech tasks.
        \label{fig:modelone}}
        \vspace{-0.2cm}
	\end{center}
\end{figure*}
}

\newcommand{\insertMLStable}{
\begin{table*}[h!]
\begin{center}
\caption{Speech recognition performance on Multilingual LibriSpeech (MLS) in terms of WER. 
Models are fine-tuned with 10h or all available labeled data (full) for each language. 
We compare to \citet{pratap2020mls} who uses all labeled data (nearly 45K hours for English). 
Results are based on a 4-gram language model, except for Polish (pl) where we report Viterbi results for all settings since performance with the provided language model data resulted in inferior accuracy, as previously reported~\citep{pratap2020mls}; we denote this with ($^*$).
\label{tab:mls}}
\begin{tabular}[h!]{llrrrrrrrrrr}
\toprule
& \#ft & en & de & nl & fr & es & it & pt & pl$^*$ & Avg. \\
\midrule
\multicolumn{2}{l}{Full labeled data (h)} & 44.7K & 2K & 1.6K & 1.1K & 918 & 247 & 161 & 104  \\
\midrule
\multicolumn{1}{l}{\it Previous work} \\
\citet{pratap2020mls} & full & \bf 5.9 & \bf 6.5 & 12.0 & \bf 5.6 & \bf 6.1 & \bf 10.5 & 19.5 & 19.4 & \bf 10.7 \\
XLSR-53 & 10h & 14.6 & 8.4 & 12.8 & 12.5 & 8.9 & 13.4 & 18.2 & 17.8 & 13.8 \\
\midrule
\multicolumn{1}{l}{\it This work} \\
\xlsrpb{0.3} & 10h & 15.9 & 9.0 & 13.5 & 12.4 & 8.1 & 13.1 & 17.0 & 13.9 & 12.8 \\
\xlsrpb{1} & 10h & 12.9 & 7.4 & \bf 11.6 & 10.2 & 7.1 & 12.0 & 15.8	& 10.5 & 10.9\\
\xlsrpb{2} & 10h & 14.0 & 7.6 & 11.8 & 10.0 & 6.9 & 12.1 & \bf 15.6 & \bf 9.8 & 11.0 \\
\bottomrule
\end{tabular}
\end{center}
\end{table*}
}

\newcommand{\insertCovostEnXMain}{
\begin{table}[t]
\centering
\captionof{table}{Speech translation: results for \englishx{} directions on CoVoST-2 in terms of BLEU.
We show detailed results for four language pairs: English-German (en-de), English-Catalan (en-ca), English-Arabic (en-ar) and English-Turkish (en-tr) as well as the average performance over all 15 directions. 
For faster experimental turnaround we do not use self-training and LM-decoding as~\citet{wang2021st} and we expect these methods to be equally applicable to \xlsrp{}. 
\autoref{app:st_full} lists detailed results for all translation directions.
\label{tab:covost_enx}}            
\begin{tabular}[t]{lrrrrr}
\toprule
& en-ca & en-ar & en-de & en-tr & Avg. (15 dir) \\
\midrule
\multicolumn{4}{l}{\it Prior work} \\
XLSR-53~\citep{conneau2021xlsr} & 29.0 & 16.5 & 23.6 & 15.3 & 23.4 \\
VP-100K~\citep{wang2021vp} & 26.1 & 14.5 & 20.8 & 13.5 & 20.9 \\
XMEF JT~\citep{li2021multilingual} & 30.9 & 18.0 & 25.8 & 17.0 & 25.1 \\
wav2vec 2.0 LV-60K~\citep{wang2021st} & 32.4 & 17.4 & 23.8 & 15.4 & - \\
 + self-training + LM~\citep{wang2021st} & \bf 35.6 & \bf 20.8 & 27.2 & \bf 18.9 & - \\
\midrule
\multicolumn{1}{l}{\it This work - monolingual pretraining} \\
wav2vec 2.0 LV-60K (720M) & 32.7 & 19.4 & 27.0 & 17.7 & 26.6 \\
\midrule
\multicolumn{1}{l}{\it This work - cross-lingual pretraining} \\
\xlsrpb{0.3} & 28.7 & 16.3 & 23.6 & 15.0 & 23.2 \\
\xlsrpb{1} & 32.1 & 19.2 & 26.2 & 17.1 & 26.0 \\
\xlsrpb{2} & 34.2 & 20.7 & \bf28.3 & 18.6 & \bf 27.8 \\
\bottomrule
\end{tabular}
\end{table}
}

\newcommand{\insertCovostXEnMain}{
\begin{table}[t]
\centering
\captionof{table}{Speech translation: results for \xenglish{} directions on CoVoST-2 in terms of average BLEU for 21 directions grouped into high/mid/low-resource labeled data directions. \autoref{app:st_full} lists detailed results for all languages.
\label{tab:covost_xen}}            
\begin{tabular}[t]{lrrrrr}
\toprule
& high & mid & low & Avg. \\
\midrule
\multicolumn{4}{l}{\it Prior work} \\
XLSR-53~\citep{conneau2021xlsr} & 30.3 & 11.1 & 3.2 & 10.3 \\
VP-100K~\citep{wang2021vp} & 27.7 & 13.2 & 4.6 & 11.1 \\
XMEF-En~\citep{li2021multilingual} & 32.4 & 16.8 & 4.0 & 12.4 \\
XMEF-X~\citep{li2021multilingual} & 34.2 & 20.2 & 5.9 & 14.7 \\
\midrule
\multicolumn{1}{l}{\it This work} \\
\xlsrpb{0.3} & 30.6 & 18.9 & 5.1 & 13.2 \\
\xlsrpb{1} & 34.3 & 25.5 & 11.7 & 19.3 \\
\xlsrpb{2} & \bf 36.1 & \bf 27.7 & \bf 15.1 & \bf 22.1 \\
\bottomrule
\end{tabular}
\end{table}
}

\newcommand{\insertCovostMultBiFT}{
\begin{table}[h!]
\centering
\captionof{table}{Speech translation: multilingual vs. bilingual fine-tuning.
Multilingually fine-tuned models are trained on labeled data from either 21 \xen{} directions or 15 \enx{} directions.
Bilingually fine-tuned models are trained on a single pair.
We show average BLEU for \xlsrpb{1} over four \enx{} directions en-(ca, ar, de, tr), and the average of high- and mid-resource directions for \xen{}. Note: results are not directly comparable to the rest of the paper and are hence denoted by ($^*$).
\label{tab:covost_multbift}
}            
\begin{tabular}[t]{lrr}
\toprule
& \xen{}$^*$ & \enx{}$^*$  \\
\midrule
Bilingual fine-tuning & 20.9 & 23.4 \\
Multilingual fine-tuning & 24.2 & 23.6 \\
\bottomrule
\end{tabular}
\end{table}
}

\newcommand{\insertCovostMBART}{
\begin{table}[h!]
\centering
\captionof{table}{Speech translation: multilingual fine-tuning performance ablation in terms BLEU. 
mBART decoder initialization primarily improves mid- and low resource directions which benefit from the labeled translation data mBART was trained on.
\label{tab:covost_mbart}}
\begin{tabular}[t]{l|r|rrrr}
\toprule
& \enx{} & \multicolumn{4}{c}{\xen{}} \\
& & High & Mid & Low & Avg \\
\midrule
\xlsrpb{1} & 26.0 & 34.3 & 25.5 & 11.7 & 19.3 \\
~~~~- mBART init & 25.1 & 32.4 & 17.7 & 3.6 & 12.4 \\
\bottomrule
\end{tabular}
\end{table}
}

\newcommand{\insertVLTable}{
\begin{table}[h!]
\centering
\captionof{table}{Language identification on VoxLingua107. We report the error rate on the development set spanning 33 languages.
\label{tab:lid_voxlingua}}            
\begin{tabular}[t]{lrrr}
\toprule
& \multicolumn{3}{c}{Error Rate (\%)} \\
& 0...5 sec & 5...20 sec & Average \\
\midrule
\multicolumn{1}{l}{\it Previous work} \\
\citet{valk2020voxlingua107} & 12.3 & 6.1 & 7.1 \\
\citet{speechbrain} & - & - & 6.7 \\
\midrule
\multicolumn{1}{l}{\it This work} \\
wav2vec 2.0 LV-60K (300M) & 11.5 & 6.3 & 7.2 \\
\xlsrpb{0.3} & \bf 9.1 & \bf 5.0 & \bf 5.7 \\
\bottomrule
\end{tabular}
\end{table}
}

\newcommand{\insertVoxCelebOneTable}{
\begin{table}[h]
\centering
\captionof{table}{Speaker identification accuracy on VoxCeleb1 in terms of accuracy. We report baselines from previous work as well as \xlsrpb{}.
\label{tab:voxceleb1}}            
\begin{tabular}[t]{lr}
\toprule
& Accuracy (\%) \\
\midrule
\multicolumn{1}{l}{\it Previous work} \\
CNN (\cite{Nagrani17}) & 80.5 \\
SUPERB-Hubert Large \citep{yang2021superb} & 90.3 \\
\midrule
\multicolumn{1}{l}{\it This work} \\
\xlsrpb{0.3} & \bf 95.8 \\
\bottomrule
\end{tabular}
\end{table}
}

\newcommand{\insertLanguageTable}{
\begin{table}[h!]
\centering
\begin{small}
\begin{tabular}[t]{llllr}
\toprule
Language & ISO & Family & Sub-grouping & Train data (h) \\
\midrule
Abkhazian & ab & Abkhaz-Adyge & Abkhaz-Abazin & \textless1 \\
Afrikaans & af & Indo-European & Germanic & 87 \\
Albanian & sq & Indo-European & Albanian & 56 \\
Amharic & am & Afro-Asiatic & Semitic & 65 \\
Arabic & ar & Afro-Asiatic & Semitic & 95 \\
Armenian & hy & Indo-European & Armenian & 55 \\
Assamese & as & Indo-European & Indo-Iranian & 179 \\
Azerbaijani & az & Turkic & Southern & 47 \\
Bashkir & ba & Turkic & Western & 47 \\
Basque & eu & Language isolate &  & 113 \\
Belarusian & be & Indo-European & Balto-Slavic & 106 \\
Bengali & bn & Indo-European & Indo-Iranian & 100 \\
Bosnian & bs & Indo-European & Balto-Slavic & 83 \\
Breton & br & Indo-European & Celtic & 42 \\
Bulgarian & bg & Indo-European & Balto-Slavic & 17616 \\
Burmese & my & Sino-Tibetan & Tibeto-Burman & 33 \\
Cantonese & yue & Sino-Tibetan & Chinese & 130 \\
Catalan & ca & Indo-European & Italic & 691 \\
Cebuano & ceb & Austronesian & Malayo-Polynesian & 42 \\
Central Khmer & km & Austro-Asiatic & Mon-Khmer & 33 \\
Chinese CN & zh & Sino-Tibetan & Chinese & 90 \\
Chinese HK & zh-HK & Sino-Tibetan & Chinese & 51 \\
Chinese TW & zh-TW & Sino-Tibetan & Chinese & 53 \\
Chuvash & cv & Turkic & Bolgar & 4 \\
Croatian & hr & Indo-European & Balto-Slavic & 2520 \\
Czech & cs & Indo-European & Balto-Slavic & 18514 \\
Danish & da & Indo-European & Germanic & 13588 \\
Divehi & dv & Indo-European & Indo-Iranian & 18 \\
Dutch & nl & Indo-European & Germanic & 20070 \\
English & en & Indo-European & Germanic & 69493 \\
Esperanto & eo & Constructed language &  & 97 \\
Estonian & et & Uralic & Coastal Finnic & 10652 \\
Faroese & fo & Indo-European & Germanic & 54 \\
Finnish & fi & Uralic & Finnic & 13981 \\
French & fr & Indo-European & Italic & 23973 \\
Galician & gl & Indo-European & Italic & 57 \\
Ganda & lg & Atlantic-Congo & Volta-Congo & 3 \\
Georgian & ka & Kartvelian & Georgian & 127 \\
German & de & Indo-European & Germanic & 25378 \\
Greek & el & Indo-European & Greek & 17761 \\
Guarani & gn & Tupian & Tupí-Guaraní & 2 \\
Gujarati & gu & Indo-European & Indo-Iranian & 37 \\
Haitian & ht & Creole & French-based & 138 \\
Hakha Chin & cnh & Sino-Tibetan & Tibeto-Burman & 2 \\
Hausa & ha & Afro-Asiatic & Chadic & 75 \\
Hawaiian & haw & Austronesian & Malayo-Polynesian & 10 \\
Hebrew & he & Afro-Asiatic & Semitic & 77 \\
Hindi & hi & Indo-European & Indo-Iranian & 65 \\
Hungarian & hu & Uralic & & 17421 \\
Icelandic & is & Indo-European & Germanic & 73 \\
Indonesian & id & Austronesian & Malayo-Polynesian & 41 \\
Interlingua & ia & Constructed language & & 9 \\
Irish & ga & Indo-European & Celtic & 3 \\
Italian & it & Indo-European & Italic & 21943 \\
Japanese & ja & Japonic & & 49 \\
Javanese & jv & Austronesian & Malayo-Polynesian & 42 \\
Kabyle & kab & Afro-Asiatic & Berber & 520 \\
Kannada & kn & Dravidian & Southern & 36 \\
Kazakh & kk & Turkic & Western & 98 \\
Kinyarwanda & rw & Atlantic-Congo & Volta-Congo & 1199 \\
\bottomrule
\end{tabular}
\end{small}
\end{table}

\begin{table}[h!]
\centering
\begin{small}
\begin{tabular}[t]{llllr}
\toprule
Language & ISO & Family & Sub-grouping & Train data (h) \\
\midrule
Kyrgyz & ky & Turkic & Western & 11 \\
Korean & ko & Koreanic & & 61 \\
Kurmanji & ku & Indo-European & Indo-Iranian & 38 \\
Lao & lo & Kra-Dai & Kam-Tai & 93 \\
Latin & la & Indo-European & Italic & 53 \\
Latvian & lv & Indo-European & Balto-Slavic & 13126 \\
Lingala & ln & Atlantic-Congo & Volta-Congo & 72 \\
Lithuanian & lt & Indo-European & Balto-Slavic & 14423 \\
Luxembourgish & lm & Indo-European & Germanic & 60 \\
Macedonian & mk & Indo-European & Balto-Slavic & 89 \\
Malagasy & mg & Austronesian & Malayo-Polynesian & 87 \\
Malay & ms & Austronesian & Malayo-Polynesian & 66 \\
Malayalam & ml & Dravidian & Southern & 38 \\
Maltese & mt & Afro-Asiatic & Semitic & 9120 \\
Manx & gv & Indo-European & Celtic & 3 \\
Maori & mi & Austronesian & Malayo-Polynesian & 27 \\
Marathi & mr & Indo-European & Indo-Iranian & 68 \\
Mongolian & mn & Mon-Khi & Mongolic & 68 \\
Nepali & ne & Indo-European & Indo-Iranian & 58 \\
Norwegian & no & Indo-European & Germanic & 85 \\
Nynorsk & nn & Indo-European & Germanic & 45 \\
Occitan & oc & Indo-European & Italic & 12 \\
Oriya & or & Indo-European & Indo-Iranian & 1 \\
Pashto & ps & Indo-European & Indo-Iranian & 109 \\
Persian & fa & Indo-European & Indo-Iranian & 321 \\
Polish & pl & Indo-European & Balto-Slavic & 20912 \\
Portuguese & pt & Indo-European & Italic & 17797 \\
Punjabi & pa & Indo-European & Indo-Iranian & 43 \\
Romanian & ro & Indo-European & Italic & 17515 \\
Romansh Sursilvan & rm & Indo-European & Italic & 6 \\
Romansh Vallader & rm & Indo-European & Italic & 2 \\
Russian & ru & Indo-European & Balto-Slavic & 166 \\
Sakha & sah & Turkic & Northern & 4 \\
Sanskrit & sa & Indo-European & Indo-Iranian & 12 \\
Scots & sco & Indo-European & Germanic & 2 \\
Serbian & sr & Indo-European & Balto-Slavic & 40 \\
Shona & sn & Atlantic-Congo & Volta-Congo & 24 \\
Sindhi & sd & Indo-European & Indo-Iranian & 67 \\
Sinhala & si & Indo-European & Indo-Iranian & 54 \\
Slovakian & sk & Indo-European & Balto-Slavic & 11925 \\
Slovenian & sl & Indo-European & Balto-Slavic & 11210 \\
Somali & so & Afro-Asiatic & Cushitic & 82 \\
Sorbian Upper & hsb & Indo-European & Balto-Slavic & 2 \\
Spanish & es & Indo-European & Italic & 22258 \\
Sundanese & su & Austronesian & Malayo-Polynesian & 51 \\
Swahili & sw & Atlantic-Congo & Volta-Congo & 91 \\
Swedish & sv & Indo-European & Germanic & 16325 \\
Tagalog & tl & Austronesian & Malayo-Polynesian & 150 \\
Tajik & tg & Indo-European & Indo-Iranian & 51 \\
Tamil & ta & Dravidian & Dravidian & 118 \\
Tatar & tt & Turkic & Western & 107 \\
Telugu & te & Dravidian & South-Central & 62 \\
Thai & th & Kra-Dai & Kam-Tai & 57 \\
Tibetan & bo & Sino-Tibetan & Tibeto-Burman & 81 \\
Tok & tpi & Creole & English-based & 36 \\
Turkish & tr & Turkic & Southern & 136 \\
Turkmen & tk & Turkic & Southern & 68 \\
Ukrainian & uk & Indo-European & Balto-Slavic & 72 \\
Urdu & ur & Indo-European & Indo-Iranian & 34 \\
Uzbek & uz & Turkic & Eastern & 36 \\
\bottomrule
\end{tabular}
\end{small}
\end{table}

\begin{table}[t]
\centering
\begin{small}
\begin{tabular}[h!]{llllr}
\toprule
Language & ISO & Family & Sub-grouping & Train data (h) \\
\midrule
Vietnamese & vi & Austro-Asiatic & Mon-Khmer & 131 \\
Votic & vot & Uralic & Finnic & \textless1 \\
Waray & war & Austronesian & Malayo-Polynesian & 9 \\
Welsh & cy & Indo-European & Celtic & 156 \\
Western Frisian & fy-NL & Indo-European & Germanic & 15 \\
Yiddish & yi & Indo-European & Germanic & 37 \\
Yoruba & yo & Atlantic-Congo & Volta-Congo & 75 \\
Zulu & zu & Atlantic-Congo & Volta-Congo & 56 \\
\bottomrule
\end{tabular}
\end{small}
\caption{Languages on which \xlsrp{} is trained on together with their ISO code, language family, sub-grouping and the amount of data for pretraining.
\label{tab:langlist}
}
\end{table}
}

\newcommand{\insertModels}{
\begin{table*}[h!]
\begin{center}
\resizebox{1\linewidth}{!}{
\begin{tabular}[b]{lrlrrrrrrr}
\toprule
Model & \#lgs & Train datasets & $B$ & $H_{m}$ & $H_{ff}$ & $A$ & \#params \\
\midrule
XLSR-53 & 53 & MLS, CV, BBL & 24 & 1024 & 4096 & 16 & 317M \\
VP-100K & 23 & VP-100K & 24 & 1024 & 4096 & 16 & 317M \\
\midrule
\xlsrpb{0.3} & 128 & VP-400K, MLS, CV, VL, BBL & 24 & 1024 & 4096 & 16 & 317M \\
\xlsrpb{1} & 128 & VP-400K, MLS, CV, VL, BBL & 48 & 1024 & 4096 & 16 & 965M \\
\xlsrpb{2} & 128 & VP-400K, MLS, CV, VL, BBL & 48 & 1920 & 7680 & 16 & 2162M \\
\bottomrule
\end{tabular}
}
\caption{\textbf{Model architectures.}
We show details for prior work, XLSR-53~\citep{conneau2021xlsr}, VP-100K~\citep{wang2021voxpopuli}, and the \xlsrp{} models: the number of languages (\#lgs), the pretraining data (Datasets), the number of blocks ($B$), the number of hidden states ($H_{m}$), the inner dimension of feed-forward blocks ($H_{ff}$), the number of attention heads (A) and the total number of parameters (\#params).
\label{tab:models}}
\end{center}
\vspace{-0.4cm}
\end{table*}
}

\newcommand{\insertCovostXEnFull}{
\begin{table}[h!]
\begin{tabular}{l|rrrr|rrrrr|rr}
\toprule
& \multicolumn{4}{c|}{High-resource} & \multicolumn{5}{c|}{Mid-resource} & \\
\midrule
\xenglish{} & fr & de & es & ca & fa & it & ru & pt & zh & tr & ar \\
Train Hours & 264h & 184h & 113h & 136h & 49h & 44h & 18h & 10h & 10h & 4h & 2h \\
\midrule 
\multicolumn{8}{l}{\it Prior work} \\
XLSR-53 & 32.3 & 26.9 & 33.3 & 28.6 & 7.1 & 26.2 & 4.9 & 10.5 & 6.9 & 3.7 & 1.2 \\
VP-100K  & 30.4 & 23.4 & 31.1 & 25.7 & 2.0 & 26.3 & 21.7 & 13.6 & 2.6 & 0.9 & 0.8 \\
XMEF-En & 35.0 & 28.2 & 35.2 & 31.1 & 3.6 & 27.6 & 22.8 & 24.1 & 6.0 & 4.8 & 2.8 \\
XMEF-X & 36.1 & 30.6 & 38.1 & 31.8 & 8.5 & 31.9 & 30.9 & 20.7 & 8.9 & 9.4 & 6.4 \\
\midrule
\multicolumn{8}{l}{\it This work} \\
\xlsrpb{0.3} & 32.9 & 26.7 & 34.1 & 28.7 & 5.9 & 29.0 & 26.4 & 28.3 & 4.9 & 4.6 & 3.0 \\
\xlsrpb{1} & 36.2 & 31.2 & 37.9 & 31.9 & 9.6  & 33.1 & 37.0 & 39.3 & 8.7 & 12.8 & 12.2 \\
\xlsrpb{2} & \bf 37.6 & \bf 33.6 & \bf 39.2 & \bf 33.8 & \bf 12.9 & \bf 34.9 & \bf 39.5 & \bf 41.8 & \bf 9.4 & \bf 16.7 & \bf 17.1 \\
\bottomrule
\end{tabular}    
\begin{tabular}{l|rrrrrrrrrr|r}
\toprule
 & \multicolumn{10}{c|}{Low-resource} & \\
\midrule
\xenglish{} & et & mn & nl & sv & lv & sl & ta & ja & id & cy & Avg \\
Train Hours & 3h & 3h & 7h & 2h & 2h & 2h & 2h & 2h & 2h & 2h	& \\
\midrule 
\multicolumn{8}{l}{\it Prior work} \\
XLSR-53 & 0.7 & 0.6 & 20.5 & 2.8 & 1.9 & 0.5 & 0.1 & 0.4 & 0.6 & 5.6 & 10.3 \\
VP-100K  & 4.6 & 0.3 & 18.3 & 11.7 & 9.0 & 8.1 & 0.1 & 0.2 & 0.7 & 0.6 & 11.1 \\
XMEF-En  & 1.5 & 0.9 & 14.2 & 5.0 & 4.9 & 5.0 & 0.8 & 1.7 & 3.7 & 2.3 & 12.4 \\
XMEF-X & 2.5 & 1.2 & 24.0 & 4.0 & 5.0 & 5.6 & \bf 0.9 & 1.0 & 2.8 & 8.1 & 14.7 \\
\midrule
\multicolumn{8}{l}{\it This work} \\
\xlsrpb{0.3} & 3.5 & 0.4 & 22.0 & 10.3 & 6.0 & 6.6 & 0.2 & 0.6 & 1.4 & 2.5 & 13.2 \\
\xlsrpb{1}   & 8.3 & 0.8 & 28.2 & 24.7 & 16.0 & 16.7 & 0.3 & 1.9 & 10.3 &  8.6 & 19.3 \\
\xlsrpb{2} & \bf 11.1 & \bf 1.6 & \bf 31.7 & \bf 29.6 & \bf 19.5 & \bf 19.6 & 0.5 & \bf 3.5 & \bf 16.5 & \bf 14.0 & \bf 22.1 \\
\bottomrule
\end{tabular}    
\caption{CoVoST-2 \xenglish{} full results. We report baseline results from  XLSR-53 \citep{conneau2021xlsr}, VP-100K \cite{wang2021voxpopuli}, XMEF-En and XMEF-X \cite{li2021multilingual}.}
\label{tabcovost_xen_full}
\end{table}
}

\newcommand{\insertCovostEnXFull}{
\begin{table}[h!]
\begin{tabular}{l|rrrrrrrr}
\toprule
\englishx{} & ca & ar & de & tr & zh & fa & et & mn \\
Train Hours & 430h & 430h & 430h & 430h & 430h & 430h & 430h & 430h \\
\midrule 
\multicolumn{8}{l}{\it Prior work} \\
XLSR-53 & 29.0 & 16.5 & 23.6 & 15.3 & 33.7 & 19.4 & 19.8 & 13.3 \\
VP-100K  & 26.1 & 14.5 & 20.8 & 13.5 & 30.7 & 17.1 & 17.4 & 12.0 \\
XMEF-JT & 30.9 & 18.0 & 25.8 & 17.0 & 33.3 & 21.5 & 22.1 & 14.8 \\
Wav2vec 2.0 (0.3B)  & 32.4 & 17.4 & 23.8 & 15.4 & - & - & - & -  \\
+ self-training + LM  & \bf 35.6 & \bf 20.8 & 27.2 & \bf 18.9 & - & - & - & -  \\
\midrule
\multicolumn{8}{l}{\it This work - monolingual pretraining (LV-60K)} \\
Wav2vec 2.0 (0.72B) & 32.7 & 19.4 & 27.0 & 17.7 & 37.4 & 21.8 & 22.9 & 15.4 \\
\midrule
\multicolumn{8}{l}{\it This work - cross-lingual pretraining} \\
\xlsrpb{0.3} & 28.7 & 16.3 & 23.6 & 15.0 & 33.5 & 19.0 & 19.6 & 13.2 \\
\xlsrpb{1} & 32.1 & 19.2 & 26.2 & 17.1 & 36.7 & 21.3 & 22.4 & 14.9 \\
\xlsrpb{2} & 34.2 & 20.7 & \bf 28.3 & 18.6 & \bf 38.5 & \bf 22.9 & \bf 24.1 & \bf 16.2 \\
\bottomrule
\end{tabular}    
\begin{tabular}{l|rrrrrrr|r}
\toprule
\englishx{} &  sv & lv & sl & ta & ja & id & cy & Avg\\
Train Hours & 430h & 430h & 430h & 430h & 430h & 430h & 430h & \\
\midrule 
\multicolumn{8}{l}{\it Prior work} \\
XLSR-53 & 29.6 & 19.3 & 22.8 & 15.8 & 37.1 & 27.6 & 28.8 & 23.4 \\
VP-100K & 26.3 & 16.7 & 20.2 & 13.9 & 33.9 & 24.9 & 26.0 & 20.9 \\
XMEF-JT & 29.6 & 21.5 & 25.1 & 17.8 & 39.3 & 29.9 & 30.6 & 25.1 \\
Wav2vec 2.0 (0.3B)  & - & - & - & - & - & - & - & -  \\
+ self-training + LM  & - & - & - & - & - & - & - & -  \\
\midrule
\multicolumn{8}{l}{\it This work - monolingual pretraining (LV-60K)} \\
Wav2vec 2.0 (0.72B) & 33.0 & 22.5 & 26.1 & 18.4 & 40.3 & 31.2 & 32.8 & 26.6 \\
\midrule
\multicolumn{8}{l}{\it This work - cross-lingual pretraining} \\
\xlsrpb{0.3} & 29.1 & 19.3 & 22.4 & 15.6 & 36.9 & 27.4 & 28.9 & 23.2 \\
\xlsrpb{1} & 32.3 & 22.0 & 25.4 & 18.1 & 39.9 & 30.3 & 31.8 & 26.0 \\
\xlsrpb{2} & \bf 34.5 & \bf 23.5 & \bf 27.6 & \bf 19.8 & \bf 41.5 & 32.5 & \bf 33.8 & \bf 27.8 \\
\bottomrule
\end{tabular}    
\caption{CoVoST-2 \englishx{} full results. We report baselines results from XLSR-53 \citep{conneau2021xlsr}, VP-100K \citep{wang2021voxpopuli}, XMEF-JT \citep{li2021multilingual} and wav2vec 2.0 LV-60K with self-training \citep{wang2021st}}
\label{tab:covost_enx_full}
\end{table}
}

\newcommand{\insertLStable}{
\begin{table*}[t]
\begin{center}
\caption{LibriSpeech ASR results in terms of WER. 
Models are fine-tuned with 10min, 1h or 10h of annotated data. 
We compare XLS-R to wav2vec 2.0~\citep{baevski2020wav} with the same number of parameters (317M). 
We do not use any language model for these experiments.
Cross-lingual training with higher capacity such as for \xlsrpb{1} obtains competitive performance.
\label{tab:ls}}
  \begin{tabular}{lrrrr}
    \toprule
    \multirow{2}{*}{Model} & \multicolumn{2}{c}{dev} & \multicolumn{2}{c}{test} \\
    & clean & other & clean & other \\
    \midrule
    \midrule
    \bfseries 10 min labeled \\
    wav2vec 2.0 LV-60K (0.3B) & 31.7 & 35.0 & 32.1 & 34.5 \\
    \xlsrpb{0.3} & 33.3 & 39.8 & 34.1 & 39.6 \\
    \xlsrpb{1} & \bf 28.4 & \bf 32.5 & \bf 29.1 & \bf 32.5 \\
    \midrule
    \midrule
    \bfseries 1h labeled \\
    wav2vec 2.0 LV-60K (0.3B) & 13.7 & \bf 16.9 & 13.7 & \bf 17.1 \\
    \xlsrpb{0.3} & 17.1 & 23.7 & 16.8 & 24.0\\
    \xlsrpb{1} & \bf 13.2 & 17.0 & \bf 13.1 & 17.2 \\
    \midrule
    \midrule
    \bfseries 10h labeled \\
    wav2vec 2.0 LV-60K (0.3B) & \bf 5.7 & \bf 9.2 & \bf 5.6 & \bf 9.4 \\
    \xlsrpb{0.3} & 8.3 & 15.1 & 8.3 & 15.4 \\
    \xlsrpb{1} & 5.9 & 10.5 & 5.9 & 10.6 \\
\bottomrule
\end{tabular}
\end{center}
\end{table*}
}

\pgfplotstableread[row sep=\\,col sep=&]{
lang & hours \\
English & 69489.4 \\
German & 25374.3 \\
French & 23969.2 \\
Spanish & 22253.7 \\
Italian & 21938.9 \\
Polish & 20907.9 \\
Dutch & 20065.8 \\
Czech & 18510.4 \\
Portuguese & 17793 \\
Greek & 17757.5 \\
Bulgarian & 17614.1 \\
Romanian & 17510.6 \\
Hungarian & 17417.4 \\
Swedish & 16320.9 \\
Lithuanian & 14420.2 \\
Finnish & 13978.7 \\
Danish & 13586.3 \\
Latvian & 13121.8 \\
Slovak & 11922.9 \\
Slovenian & 11205.9 \\
Estonian & 10648.5 \\
Maltese & 9116.02 \\
Croatian & 2518.27 \\
Kinyarwanda & 1196.95 \\
Catalan & 689.05 \\
Kabyle & 518.47 \\
Persian & 319.51 \\
Assamese & 179 \\
Russian & 164.44 \\
Welsh & 153.66 \\
Tagalog & 150.13 \\
Haitian & 137.71 \\
Turkish & 133.81 \\
Vietnamese & 130.71 \\
Cantonese & 130.12 \\
Georgian & 126.55 \\
Tamil & 116.17 \\
Basque & 111.24 \\
Belarusian & 106.49 \\
Tatar & 104.66 \\
Bengali & 99.98 \\
Kazakh & 98.49 \\
Esperanto & 95.17 \\
Arabic & 93.5 \\
Lao & 92.56 \\
Swahili & 91.3 \\
Macedonian & 89.1 \\
Chinese_CN & 88.42 \\
Malagasy & 87.01 \\
Afrikaans & 86.7 \\
Norwegian & 85.41 \\
Bosnian & 83.52 \\
Somali & 82.24 \\
Tibetan & 80.82 \\
Hebrew & 77.03 \\
Yoruba & 74.91 \\
Hausa & 74.85 \\
Icelandic & 73.23 \\
Lingala & 72.07 \\
Pashto & 70.68 \\
Ukrainian & 70.19 \\
Marathi & 67.92 \\
Turkmen & 67.62 \\
Sindhi & 67 \\
Malay & 66.3 \\
Mongolian & 65.97 \\
Hindi & 65.02 \\
Amharic & 64.87 \\
Telugu & 61.87 \\
Korean & 61.35 \\
Luxembourgish & 59.94 \\
Nepali & 57.79 \\
Galician & 57.27 \\
Albanian & 56.53 \\
Zulu & 56.41 \\
Armenian & 55.21 \\
Thai & 55.06 \\
Faroese & 53.84 \\
Sinhala & 53.69 \\
Latin & 53.53 \\
Chinese_TW & 51.25 \\
Tajik & 51.04 \\
Sundanese & 50.64 \\
Chinese_HK & 48.66 \\
Japanese & 47.87 \\
Bashkir & 46.91 \\
Azerbaijani & 46.7 \\
Norwegian_Nynorsk & 45.06 \\
Punjabi & 43.06 \\
Cebuano & 42.23 \\
Javanese & 42.03 \\
Breton & 40.32 \\
Serbian & 39.85 \\
Indonesian & 39.2 \\
Kurmanji & 38.04 \\
Malayalam & 38.02 \\
Pushto & 37.57 \\
Gujarati & 36.9 \\
Yiddish & 36.66 \\
Kannada & 36.42 \\
Uzbek & 36.36 \\
Tok & 35.68 \\
Urdu & 34.1 \\
Burmese & 32.9 \\
Central_Khmer & 32.68 \\
Maori & 27.13 \\
Shona & 24.32 \\
Divehi & 16.07 \\
Dutch_fy_NL & 12.74 \\
Occitan & 12.23 \\
Sanskrit & 12.22 \\
Hawaiian & 9.76 \\
Kirghiz & 8.85 \\
Waray & 8.76 \\
Abkhazian & 8.23 \\
Interlingua & 6.81 \\
Romansh_sursilv & 3.69 \\
Chuvash & 3.46 \\
Sakha & 3.4 \\
Manx & 3.38 \\
Irish & 2.54 \\
Ganda & 2.47 \\
Scots & 2.34 \\
Guarani & 2.03 \\
Sorbian_Upper & 2 \\
Hakha_Chin & 1.88 \\
Romansh_vallader & 1.8 \\
Oriya & 0.72 \\
Votic & 0 \\
}\langdata

\newcommand*\samethanks[1][\value{footnote}]{\footnotemark[#1]}
\newcommand{\xlsrp}{XLS-R}
\newcommand{\xlsrpb}[1]{\xlsrp{} {(#1B)}}

\newcommand{\enx}{En $\rightarrow$ X}
\newcommand{\englishx}{English $\rightarrow$ X}
\newcommand{\xen}{X $\rightarrow$ En}
\newcommand{\xenglish}{X $\rightarrow$ English}

\newcommand{\Inp}{\mathcal{X}}
\newcommand{\Feat}{\mathcal{Z}}
\newcommand{\QFeat}{\mathcal{Q}}

\newcommand{\Context}{\mathcal{C}}

\newcommand{\ze}{z}
\newcommand{\zq}{q}
\newcommand{\zqt}{\tilde{q}}

\newcommand{\cc}{c}

\iclrfinalcopy 
\begin{document}

\maketitle

\begin{abstract}
This paper presents \xlsrp{}, a large-scale model for cross-lingual speech representation learning based on wav2vec 2.0. 
We train models with up to 2B parameters on nearly half a million hours of publicly available speech audio in 128 languages, an order of magnitude more public data than the largest known prior work.
Our evaluation covers a wide range of tasks, domains, data regimes and languages, both high and low-resource. 
On the CoVoST-2 speech translation benchmark, we improve the previous state of the art by an average of 7.4 BLEU over 21 translation directions into English.
For speech recognition, \xlsrp{} improves over the best known prior work on BABEL, MLS, CommonVoice as well as VoxPopuli, lowering error rates by 14-34\% relative on average.
\xlsrp{} also sets a new state of the art on VoxLingua107 language identification.
Moreover, we show that with sufficient model size, cross-lingual pretraining can perform as well as English-only pretraining when translating English speech into other languages, a setting which favors monolingual pretraining.
We hope \xlsrp{} can help to improve speech processing tasks for many more languages of the world. 
Models and code are available at {\small\url{www.github.com/pytorch/fairseq/tree/master/examples/wav2vec/xlsr}}.\footnote{Hugging Face: {\small \url{https://huggingface.co/models?other=xls_r}}}
\end{abstract}

\section{Introduction}
Self-supervised learning of generic neural representations has gathered much recent interest with a large body of work in natural language processing (NLP; ~\citealt{radford2018improving,baevski2019cloze,devlin2018bert,raffel2019exploring}), computer vision~\citep{chen2020simple,he2020momentum,caron2021emerging} as well as speech processing~\citep{oord2018cpc,schneider2019wav2vec,baevski2020wav,hsu2020hubert,chung2021w2vbert}. 
Self-supervised learning provides general representations that can be used across domains and languages. 

Multilingually pretrained NLP models such as mBERT~\citep{devlin2018bert}, XLM-R~\citep{conneau-etal-2020-unsupervised} or mT5~\citep{xue2020mt5} brought significant improvements in multilingual language understanding~\citep{conneau2018xnli,hu2020xtreme,ruder2021xtreme}.
These models offer a promising path towards more ubiquitous NLP technology by improving performance for low-resource languages through leveraging data from high-resource languages.
Furthermore, it is only necessary to maintain a single multilingual model instead of a myriad of monolingual models.

For speech processing, self-supervised approaches such as wav2vec 2.0~\citep{baevski2020wav,xu2021self} have also been extended to the multilingual setting~\citep{kawakami2020learning,conneau2021xlsr}. 
The recent XLSR~\citep{conneau2021xlsr} leverages cross-lingual transfer from high-resource languages to build better representations for languages with little unlabeled data.
The largest model, XLSR-53, was trained on about 50K hours of public training data in 53 languages and comprises about 300M parameters~\citep{conneau2021xlsr}.
But such models only scratch the surface of self-supervised cross-lingual speech representation learning.

In natural language processing, language models are trained on very large datasets, spanning billions of documents such as CC100~\citep{wenzek2020ccnet} or mC4~\citep{xue2020mt5} to fit models with tens of billions and even trillions of parameters~\citep{brown2020gpt3,goyal2021xlmr,fedus2021switch} with strong results on established benchmarks. 
In contrast, scaling efforts in speech have focused either on supervised multilingual models~\citep{li2021scaling} or monolingual self-supervised models, counting a billion or more parameters~\citep{zhang2020pushing,zhang2021bigssl}, while cross-lingually pretrained speech models are much smaller in scale.

To this end, we present \xlsrp{}, a large-scale cross-lingually pretrained wav2vec 2.0 model (see illustration in \autoref{fig:modelone}) whose name is inspired by XLM-R in NLP.
It leverages new publicly available VoxPopuli data, comprising 372K hours of unannotated speech~\citep{wang2021voxpopuli}, the MLS corpus~\citep{pratap2020mls}, CommonVoice~\citep{ardila2019common}, BABEL~\citep{gales2014babel} and VoxLingua107~\citep{valk2020voxlingua107} to cover 128 different languages from various regions of the world. 
To our knowledge, this is the largest effort to date, in making speech technology accessible for many more languages using publicly available data.

\insertXLSR


\section{Background}
\label{sec:background}

Our work builds on~\citet{conneau2021xlsr} who pretrain wav2vec 2.0 models on data from multiple languages.
wav2vec 2.0 contains a convolutional feature encoder $f: \Inp \mapsto \Feat$ to map raw audio~$\Inp$ to latent speech representations $\ze_1, \dots, \ze_T$ which are input to a Transformer $g: \Feat \mapsto \Context$ to output context representations $\cc_1, \dots, \cc_T$~\citep{baevski2019vqwav2vec}.
Each $\ze_t$ represents 25ms of audio strided by 20ms and the Transformer architecture follows BERT~\citep{vaswani2017transformer,devlin2018bert}.

During training, feature encoder representations are discretized to $\zq_1, \dots, \zq_T$ with a quantization module $\Feat \mapsto \QFeat$ to represent the targets in the objective.
The quantization module uses a Gumbel softmax to choose entries form the codebooks and the chosen entries are concatenated to obtain $\zq$~\citep{jegou2011ieee,jang2016gumbel,baevski2019vqwav2vec}.

The model is trained by solving a contrastive task over masked feature encoder outputs.
At training time, spans of ten time steps with random starting indices are masked.
The objective requires identifying the true quantized latent $\zq_t$ for a masked time-step within a set of $K=100$ distractors $\mathbf{Q}_t$ sampled from other masked time steps.
$-\log \frac{\exp(sim(\cc_t, \zq_t))}{\sum_{\zqt \sim \mathbf{Q}_t} \exp(sim(\cc_t, \zqt))}$
where $\cc_t$ is the output of the Transformer, and $sim(\mathbf{a}, \mathbf{b})$ denotes cosine similarity.
The objective is augmented by a codebook diversity penalty to encourage the model to use all codebook entries~\citep{dieleman2018challenge}.


The model is trained on multiple languages to obtain cross-lingual representations.
Specifically, training batches contain samples from multiple languages $L$~\citep{devlin2018bert,lample2019xlm,conneau2021xlsr} by sampling from a distribution $p_l\sim\left(\frac{n_l}{N}\right)^{\alpha}$ where $l=1,\ldots,L$, while $n_l$ is the amount of unlabeled data for each language, and $\alpha$ is the upsampling factor which controls the trade-off between high- and low-resource languages during pretraining.

\section{Data and Evaluation}
In this section, we outline the datasets on which \xlsrp{} is pretrained as well as the language coverage. We also describe the downstream tasks on which we evaluate our models.

\subsection{Training Data}
\label{sec:traindata}
We pretrain our models on a total of 436K hours of publicly available data from the following sources:

\begin{itemize}
    \item \textbf{VoxPopuli (VP-400K)} comprises a total of 372K hours of data\footnote{There are 400K hours before removing leading and trailing silences.} in 23 European languages of parliamentary speech from the European parliament~\citep{wang2021voxpopuli}. This makes it the largest publicly available speech corpus for semi-supervised learning. 
    \item \textbf{Multilingual Librispeech (MLS)} contains data in eight European languages totaling around 50K hours of data~\citep{pratap2020mls}. The majority of the data is English (44K hours).
    \item \textbf{CommonVoice (CV)} is a corpus of read speech. We use the December 2020 release (v6.1;~\citealt{ardila2019common}) which covers 60 languages and over 7K hours of speech audio, ranging from over 1.6K hours for English to less than one hour for languages such as Hindi.
    \item \textbf{VoxLingua107 (VL)} is a dataset of 6.6K hours of data in 107 languages based on YouTube content~\citep{valk2020voxlingua107} with an average of 62 hours of data per language.
    \item \textbf{BABEL (BBL)} is a multilingual corpus of conversational telephone speech of about 1K hours of data in 17 African and Asian languages~\citep{gales2014babel}. 
\end{itemize}

\begin{figure}[t]
\begin{tikzpicture}
\begin{axis}[
width=1.0*\textwidth,
height=.38\textwidth,
xticklabels={,,},
ymode=log,
xmajorticks=false,
yminorticks=false,
nodes near coords align={vertical},
xmin=1,
xmax=128,
ymin=1,
ymax=100000,
ylabel={Hours of training data},
ymajorgrids=true,
xlabel={High resource $\leftarrow$ (English, German, ...)  \hspace{0.4in} Low resource $\rightarrow$ (Sinhala, Zulu, ...)},
x label style={at={(axis description cs:0.5,0.05)},anchor=north},
yticklabels={1,10,100,1000,10000,100000}
]
\addplot[black,fill=orange,opacity=0.25,smooth] table[x expr=\coordindex,y=hours]{\langdata} \closedcycle;
\end{axis}
\end{tikzpicture}
\caption{Illustration of the unlabeled training data distribution across the 128 languages of \xlsrp{}.}
\label{fig:langstats}
\end{figure}
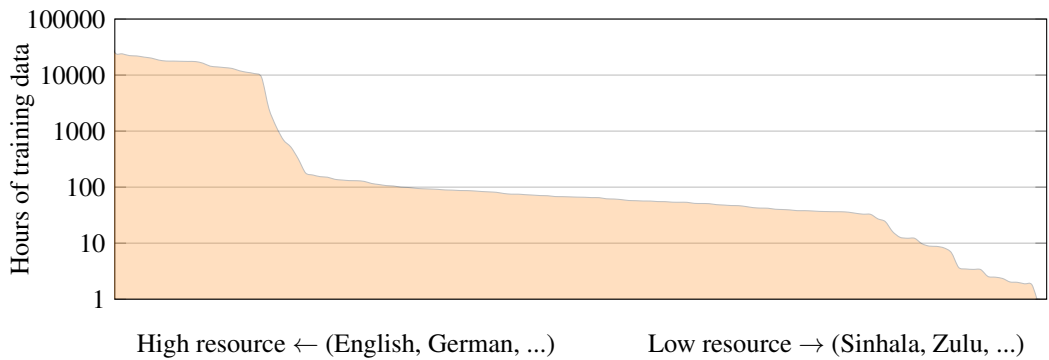

To the best of our knowledge, this is the largest dataset used for training a publicly available self-supervised speech model to date.
\autoref{fig:langstats} shows the data distribution across the 128 different languages in our training dataset.
There are about 24 high-resource languages with more than 1K hours of data each, almost all of which are European, except for Kinyarwanda which is African.
Then there is a small number of 17 mid-resource languages which have more than 100 hours of data (but less than 1K hours) which includes Catalan, Persian, Turkish, Russian, and Basque. 
Finally, the remaining 88 languages are low-resource and have less than 100 hours of data each.
\autoref{tab:langlist} lists all the languages, together with their ISO code, language family, sub-grouping and the amount of training data.

\insertLanguageTable

\subsection{Downstream Evaluation}
We evaluate on a broad and diverse set of downstream tasks to showcase the generalization ability of our pretrained models across tasks, data regimes, domains and languages.

\subsubsection{Automatic Speech Translation (AST)}
\label{sec:setup_ast}

\paragraph{CoVoST-2 AST.}{For speech translation evaluation we adopt CoVoST-2~\citep{wang2020covost2}, a multilingual speech translation benchmark based on CommonVoice~\citep{ardila2019common}.\footnote{\small\url{https://github.com/facebookresearch/covost}}
It provides data for translating from English into 15 languages (\enx{}) and from 21 languages into English (\xen{}).
The \enx{} languages are: Arabic (ar), Catalan (ca), Welsh (cy), German (de), Estonian (et), Persian (fa), Indonesian (id), Japanese (ja), Latvian (lv), Mongolian (mn), Slovenian (sl), Swedish (sv), Tamil
(ta), Turkish (tr), Chinese (zh) where each direction comprises about 430 hours of training data.
The \xen{} languages include all target languages of \enx{} as well as Spanish (es), French (fr),
Italian (it), Dutch (nl), Portuguese (pt).
We group the latter into high-resource (136-264h of train data; fr, de, es, ca), mid-resource (10-49h of train data; fa, it, ru, pt, zh), and low-resource (2-7h of train data; tr, ar, et , mn, nl, sv, lv, sl, ta, ja, id, cy) for ease of presentation.
}

\subsubsection{Automatic Speech Recognition (ASR)}
\label{sec:setup_asr}

\paragraph{BABEL ASR.}{BABEL is a challenging speech recognition benchmark from IARPA consisting of noisy telephone conversational data.\footnote{\small\url{https://catalog.ldc.upenn.edu/byyear} \tiny{LDC2016S06, LDC2016S13, LDC2017S05, LDC2017S08, LDC2016S12}} 
We evaluate on five languages: Assamese (as), Tagalog (tl), Swahili (sw), Lao (lo), and Georgian (ka). Training sets comprise between 30 and 76 hours of annotated data. Following \cite{conneau2021xlsr}, we use 10\% of the training set for validation, and report test results on the BABEL dev set. We report word error rate (WER) and use n-gram language models trained on CommonCrawl data.
}

\paragraph{Multilingual LibriSpeech ASR.}{Multilingual LibriSpeech (MLS; \citealt{pratap2020mls}) is a large corpus derived from read audiobooks of Librivox and consists of eight European languages: Dutch (nl), English (en), French (fr), German (de), Italian (it), Polish (pl), Portuguese (pt), Spanish (es).\footnote{\small\url{https://github.com/flashlight/wav2letter/blob/main/recipes/mls}} Training sets comprise between 104 hours for Polish and 44.7K hours for English. 
We use the 10 hour training splits of \cite{conneau2021xlsr} and report word error rate with the official n-gram models provided by the MLS dataset.
}

\paragraph{CommonVoice ASR.}{
Following~\cite{rivire2020unsupervised}, we use ten languages of CommonVoice for ASR evaluation: Spanish (es), French (fr), Italian (it), Kyrgyz (ky), Dutch (nl), Russian (ru), Swedish (sv), Turkish (tr), Tatar (tt) and Chinese-Hong Kong (zh-HK).\footnote{\small\url{https://dl.fbaipublicfiles.com/cpc_audio/common_voices_splits.tar.gz}}
CommonVoice contains read speech primarily from Wikipedia sentences. Following prior work~\citep{rivire2020unsupervised,conneau2021xlsr}, we fine-tune models on just one hour of labeled data per language, a few-shot scenario. Results are reported in terms of phoneme error rate (PER) without a language model.
}

\paragraph{VoxPopuli ASR.}{
Following~\cite{wang2021voxpopuli}, we evaluate on languages which have at least ten hours of labeled data which is a total of 14 languages: English (en), German (de), Italian (it), French (fr), Spanish (es), Polish (pl), Romanian (ro), Hungarian (hu), Dutch (nl), Czech (cs), Slovenian (sl), Finnish (fi), Croatian (hr), Slovakian (sk).\footnote{\small\url{https://github.com/facebookresearch/voxpopuli}}
Models are fine-tuned on the full train set, which ranges from 543 hours (English) to ten hours (Slovenian). We report word error rate without language models.
}

\paragraph{LibriSpeech ASR.}{
LibriSpeech is a widely-used evaluation benchmark for speech recognition research~\citep{panayotov2015librispeech}. It consists of 960 hours of English annotated data. Following \cite{baevski2020wav}, we use the 10 minute, 1 hour and 10 hour training splits. We compare the performance of XLS-R models against English-only wav2vec 2.0 models. We report word error rate without language models.
}

\subsubsection{Speech classification (LID and Speaker ID)}
\label{sec:setup_lid}

\paragraph{VoxLingua107 LangID.}{For spoken language identification we consider VoxLingua107~\citep{valk2020voxlingua107} which spans 107 languages.\footnote{\url{http://bark.phon.ioc.ee/voxlingua107/}} 
It consists of short speech segments automatically extracted from YouTube videos. 
The total amount of speech data in the training set is 6,628 hours and the average per language is 62 hours. 
We report accuracy on the official test set which covers 33 different languages.}

\paragraph{VoxCeleb1 SpeakerID.}{We use VoxCeleb1 for speaker identification~\citep{Nagrani17}.\footnote{\url{https://www.robots.ox.ac.uk/~vgg/data/voxceleb/vox1.html}} VoxCeleb1 is an audio-visual dataset consisting of short clips of human speech, extracted from interview videos uploaded to YouTube. 
It consists of 1,251 unique speakers and 153K utterances. 
We use the official dataset splits.
}

\section{Experimental Setup}
In this section, we give more details on architectures and hyperparameters used during pretraining and finetuning.
\subsection{Pretraining}
We use the wav2vec 2.0 implementation available in fairseq~\citep{ott2019fairseq} and evaluate several model architectures detailed in~\autoref{tab:models}.
We consider models with between 0.3B parameters to 2B parameters.
To optimize GPU memory usage, we use a fully sharded backend~\citep{rajbhandari2021zero} as well as activation checkpointing~\citep{chen20216checkpointing} as implemented in FairScale~\citep{FairScale2021}. 

Models are optimized with Adam~\citep{kingma2015adam} and the learning rate is warmed up for the first 32K steps followed by polynomial decay to zero for the remainder of training. 
Training audio sequences are cropped to a maximum of 320K samples, or 20 seconds and all models were pretrained for a total of one million updates.
\xlsrpb{0.3} was trained on 128 GPUs with nearly 2M samples on each GPU, totaling about 4.3h of data in a batch.
Larger models were trained on 200 GPUs with 800K to 1M samples on each GPU giving an effective batch size of about 2.8-3.6 hours.

Our training data covers 128 languages and five training corpora with different characteristics.
To balance data from the different languages and corpora we upsample both training corpora and languages. 
We first upsample the languages within a particular corpus using the strategy outlined in~\textsection\ref{sec:background} and then balance the different corpora using the same strategy by treating each corpus as a different language.
We use $\alpha=0.5$ in all cases.

\insertModels

\subsection{Speech translation}

To build speech translation models, we multilingually fine-tune \xlsrp{} models by training on the combined labeled data of all \enx{} or \xen{} language directions without upsampling any direction.
We stack a decoder network on top of \xlsrp{} which is a Transformer network with 12 layers, embedding size 1024, 16 attention heads and feed forward network dimension 4096.
The decoder network is initialized with weights from multilingually fine-tuned mBART~\citep{liu2020mbart,li2021multilingual,tang2021mult} and uses the same vocabulary with 250K types.
The total size of the decoder network is 459M parameters. 

In our ablations, we also consider bilingually fine-tuned speech translation models which use a much smaller decoder of 16M parameters which has seven layers, embedding size 256, 4 attention heads and feed forward network dimension 2048.
For this, a 10K byte-pair encoding (BPE;~\citealt{sennrich2016bpe}) vocabulary is built on the CoVoST 2 target text for each target language. 

We fine-tune with Adam~\citep{kingma2015adam}, a learning rate of 3e-4, label smoothing with probability 0.1, an effective batch size of 66M samples, or nearly 68 minutes, layer drop 0.05, a masking strategy similar to wav2vec 2.0 with mask length 5 and mask probability 0.15.
During fine-tuning, the wav2vec 2.0 encoder is not updated for the first 10K updates. 
Models are fine-tuned for 250K updates in total and the best checkpoint is selected based on validation BLEU. 
We choose the learning rate by searching in the interval $[3e-5, 3e-4]$.
Translations are generated with a beam size of 5.

\subsection{Automatic Speech Recognition}

For fine-tuning, we follow the settings of~\citet{baevski2020wav} by adding a linear layer on top of the pretrained model to predict the output vocabulary and train using Connectionist Temporal Classification (CTC; \citealt{graves2006ctc}).
The output vocabulary is characters for all benchmarks, except for CommonVoice where we use phonemes.
We fine-tune using Adam and the learning rate is warmed up for the first 10\% of total updates, kept constant for the next 40\% and then decayed to zero in the remaining 50\% of updates. 
Since the amount of labeled data differs widely for each dataset we found the following number of training updates to be effective: 20K updates for BABEL, 13K updates for CommonVoice, 20K updates for MLS and 50K updates for VoxPopuli. 

We found large batch sizes to be very effective.
For \xlsrpb{0.3} and \xlsrpb{1}, we use an effective batch size of 0.44 hours and for \xlsrpb{2} we used 1.06 hours. 
Learning rate as well as batch size was tuned based on dev error rate and we searched the range $[2e-5, 3e-4]$ for \xlsrpb{0.3} and \xlsrpb{1} as well as $[3e-6, 3e-5]$ for \xlsrpb{2}.
To reduce overfitting for \xlsrpb{2}, we increase stochastic depth to 15\% compared to 10\% for all other setups~\citep{huang2016stochasticdepth,fan2019layerdrop}. 
We use a masking probability of between 30-75\%, depending on the setup and which is determined on the development set.

For MLS and BABEL, we use a language model for decoding.
We tune the language model weight within the interval $[0, 5]$ and a word insertion penalty within the interval $[-5, 5]$ using Bayesian optimization.\footnote{https://github.com/facebook/Ax} 
We run 128 trials with beam 500 and choose the best set of parameters according to the dev error rate. 
For Common Voice and Vox Populi we do not use a language model.

\section{Results}
Next, we analyze the results of our pretrained models on all downstream tasks.

\subsection{Speech Translation}
We conduct an extensive study on the CoVoST-2 speech translation benchmark. 
The task entails translating speech audio in one language into another language with text as output. 
Performance is evaluated in terms of BLEU~\citep{papineni2002bleu}.
Models are simultaneously fine-tuned either on all 21 translation directions with English as target language (\xen{}) or on all the 15 directions where English is the input language (\enx{}; see~\textsection\ref{sec:setup_ast}), resulting in only two models instead of 36.

\subsubsection{\xenglish{}}
\label{sec:st_xen}

For \xenglish{} directions we group languages into high-resource, mid-resource and low-resource directions (\textsection\ref{sec:setup_ast}) and compare to several baselines: in order to directly compare to XLSR-53~\citep{conneau2021xlsr}, and VP-100K~\citep{wang2021voxpopuli}, we fine-tune these publicly available models following the same protocol as \xlsrp{}. 
We also compare to~\citet{li2021multilingual}, the best known results from the literature who either use an English-pretrained wav2vec 2.0 model (XMEF-En) for \enx{} directions or the multilingually pretrained XLSR-53 (XMEF-X) for \xen{} directions.

\autoref{tab:covost_xen} shows a new state of the art with \xlsrpb{2}, improving over the previous best result~\citep{li2021multilingual}, by 7.4 BLEU on average over all 21 directions (14.7 BLEU vs. 22.1 BLEU).
This is largely due to improvements on mid-resource (+7.5 BLEU) and low-resource (+9.2 BLEU) language directions.
\textbf{Model capacity has a large impact}: \xlsrpb{1} improves over \xlsrpb{0.3} by an average of 6.1 BLEU and \xlsrpb{2} improves by an average of 2.8 BLEU compared to \xlsrpb{1}.
Appendix \ref{app:st_full} shows detailed results for all translation directions. 

There is a trend of \textbf{larger capacity in pretrained models enabling few-shot learning for speech translation}, similar to wav2vec 2.0 enabling few-shot speech recognition~\citep{baevski2020wav,xu2020selftraining}. 
For example, on language pairs with only two hours of labeled speech translation data, \xlsrpb{2} improves over \xlsrpb{0.3} as follows: from 10.3 BLEU to 29.6 BLEU on Swedish-English, from 1.4 BLEU to 16.5 BLEU on Indonesian-English and from 3.0 BLEU to 17.1 BLEU on Arabic-English (see Appendix~\ref{app:st_full}).

\insertCovostXEnMain

\subsubsection{\englishx{}}
\label{sec:st_enx}

For \englishx{} directions we compare to previous cross-lingually pretrained models (XLSR-53, VP-100K) as well as baselines with English-only pretraining: XMEF JT, the best performing setup of~\citet{li2021multilingual} for \enx{} directions as well as wav2vec 2.0 pre-trained on 60K hours of English Libri-light data and fine-tuned following the same protocol as \xlsrp{}~\citep{kahn2020librilight,baevski2020wav}. 
The latter has the advantage of being pre-trained on exactly the same language as the input data for all translation directions while cross-lingually pretrained models need to be able to represent many different languages which puts them at a disadvantage.

\autoref{tab:covost_enx} shows that XLSR-53 now performs similarly to \xlsrpb{0.3} while for \xenglish{} \xlsrpb{0.3} performed much better (see \textsection\ref{sec:st_xen}). 
This is likely because English data dominates the training corpus of XLSR-53 which is not the case for \xlsrp{} (\textsection\ref{sec:traindata}).
Both \xlsrpb{1} and \xlsrpb{2} outperform XMEF JT showing that larger capacity results in better performance.

We also compare to prior work using the English-only pretrained wav2vec 2.0 LV-60K model~\citep{wang2021st} which additionally uses self-training and a language model for decoding. We do not use these techniques. 
Their results represent the state of the art on these four directions.
\citet{wang2021st} achieves an average BLEU of 25.6 on the four directions while as \xlsrpb{2} rivals this at an average BLEU of 25.5.
We note that self-training and LM decoding methods are equally applicable to our approach.

\xlsrpb{2} also performs well compared to English-only pretraining at 27.8 average BLEU compared to 26.6 BLEU for a wav2vec 2.0 model pretrained on 60K hours of Libri-light data and 720M parameters. 
This confirms that \textbf{with sufficient capacity, cross-lingual pretraining can perform as well as strong monolingual models}~\citep{conneau2021xlsr}.

\insertCovostEnXMain

\subsubsection{Ablations}

We build speech translation models by adopting two design decision from \citet{li2021multilingual}: multilingual fine-tuning of pretrained models on labeled speech translation data in multiple translation directions and initializing the decoder network with mBART~\citep{li2021multilingual,tang2021mult}. 
In the following we ablate these two choices to better understand their impact.

We first compare multilingual fine-tuning to bilingual fine-tuning.
For faster experimental turn-around we consider a reduced setup of four \englishx{} language directions (en-ca, en-ar, en-de, en-tr) as well as all high-resource and mid-resource \xenglish{} directions.\footnote{We also did not use mBART initialization for this ablation.}
We compare bilingual fine-tuning to models fine-tuned on all 15 \englishx{} or all 21 \xenglish{} directions.

\autoref{tab:covost_multbift} shows that \textbf{multilingual fine-tuning is particularly effective for \xenglish{} directions} where the average improvement is 3.3 BLEU (20.9 BLEU to 24.2 BLEU).
The amount of labeled data ranges from 264 hours for French $\rightarrow$ English to 10 hours for Chinese $\rightarrow$ English and multilingual fine-tuning leverages supervision from high-resource languages to improve performance for languages with less labeled data.
\textbf{Languages with less data benefit both from cross-lingual transfer during pretraining}, through training on unlabeled data in other languages, \textbf{and fine-tuning}, through labeled data from other languages~\citep{arivazhagan2019massively,conneau2021xlsr}.
For \englishx{}, multilingual fine-tuning performs roughly on par to bilingual fine-tuning which is a desirable outcome given that transfer between language directions is diminished. 
This is supported by the larger size of the decoder network in multilingual fine-tuning (\textsection\ref{sec:setup_ast}).

\insertCovostMultBiFT

Next, we analyze the impact of initializing the decoder network with mBART which was pretrained on additional labeled text-to-text machine translation data~\citep{liu2020mbart,li2021multilingual,tang2021mult}. Specifically, we use MBART-ML50N1 (49 languages to English) for \xenglish{} directions and MBART-ML501N (English to 49 languages) for \englishx{} directions.\footnote{\url{https://github.com/pytorch/fairseq/tree/main/examples/multilingual\#mbart50-models}}
We observe that \textbf{mBART initialization has little impact on \englishx{} but it leads to large improvements for \xenglish{}, especially on mid- and low-resource language directions}. 

\insertCovostMBART

Initializing the decoder network with mBART resulted in some low-resource languages moving from 1-3 BLEU to 10+ BLEU. 
The labeled translation data used to train mBART helps speech translation to adapt faster to the low supervision in mid/low-resource language pairs of the CoVoST-2 benchmarks where many language directions have only a few hours of labeled data. 
This shows that pretraining both the encoder and decoder, multilingual fine-tuning, as well as the use of extra machine translation data through mBART, enables few-shot learning for some speech translation directions which have only a few hours of labeled data.

\subsection{Speech Recognition}
Our experiments cover four common speech recognition benchmarks, 26 different languages, three different domains and both low-data and high-data regimes.
The BABEL dataset evaluates models on noisy speech (\textsection\ref{sec:res_babel}), CommonVoice presents a few-shot setup with just one hour of labeled data per language (\textsection\ref{sec:res_cv}), MLS contains read speech in multiple European languages (\textsection\ref{sec:res_mls}), and VoxPopuli contains parliamentary speech with varying amounts of labeled data (\textsection\ref{sec:res_vp}).

\subsubsection{BABEL}
\label{sec:res_babel}

BABEL consists of the hardest speech recognition setting among our four benchmarks which results in higher word error rates. 
Languages are low-resource, the speech is very noisy and corresponds to natural telephone conversation. 
Many competitions have tackled this dataset~\citep{alumae2016georgian,ragni2018confidence,inaguma2019transfer} and baselines are thus well tuned. 
We compare to the best numbers we have found in the literature, as well as our own best baselines.

Table~\ref{tab:bbl_sota} shows that \xlsrpb{0.3} outperforms the equally sized XLSR-53, which was the previous state of the art on all languages by an average of 1.4 WER, e.g.,
on Assamese (as), WER decreases from 44.1 to 42.9, on Swahili (sw) WER decreases from 26.5 to 24.3 and on Georgian (ka) WER drops from 31.1 to 28.0 WER.
XLSR-53 and \xlsrp{} were both pretrained on the same BABEL data, and the better performance of \xlsrpb{0.3} shows that pretraining on additional out-of-domain datasets such as VoxPopuli does help performance on BABEL. 
This is similar to findings for monolingual pretraining~\citep{hsu2021robust}.

Using additional capacity, \xlsrpb{1} outperforms \xlsrpb{0.3} by 2.5 WER on average.
On Georgian (ka), this corresponds to improvements of 6 WER and 7.1 WER compared to \citet{conneau2021xlsr} and \citet{alumae2016georgian}, respectively. 
Compared to results from only three years ago from \citet{ragni2018confidence} and \cite{inaguma2019transfer}, \xlsrpb{1} reduces word error rate by more than 10 WER, from 40.6 to 30.6 on Tagalog and from 35.5 to 21.2 on Lao. 
\xlsrpb{2} improves over \xlsrpb{1} by 0.8 BLEU on average showing that additional capacity can further improve performance.
\insertBABELtable

\subsubsection{CommonVoice}
\label{sec:res_cv}

CommonVoice is an easier task than BABEL because it is read-speech but we use a reduced labeled data setup which introduces a different challenge. 
Specifically, we use the train/dev/test splits of~\citet{rivire2020unsupervised} which corresponds to a few-shot scenario where only one hour of training data is available per language. 

On English speech recognition, pretraining has been shown to be particularly beneficial for low labeled data settings~\citep{baevski2020wav}.
This is similar to cross-lingual pretraining~\citep{conneau2021xlsr} where pretraining on the large MLS corpus significantly improved performance over pretraining only on CommonVoice data, e.g., on Dutch accuracy improved from 14 PER to 5.8 PER.

\autoref{tab:cv_results} shows that the additional training data of \xlsrp{} compared to XLSR-53 results in better performance of 1.1 PER on average for \xlsrpb{0.3}. 
\xlsrp{} uses the same training data as XLSR-53 plus the very large VP-400K corpus of parliamentary speech as well as the much smaller VoxLingua-107 which consists of YouTube data, both of which are out of domain with respect to the read audiobook domain of CommonVoice. 
This confirms that pretraining on more out of domain data can still improve performance~\citep{hsu2021robust}.

Furthermore, accuracy improves even on languages for which \xlsrp{} does not add any pretraining data compared to XLSR-53, e.g., Kyrgyz (ky) improves from 6.1 PER to 5.1 PER for \xlsrpb{0.3} and 4.1 PER for \xlsrpb{1} and both models are pretrained on only about 11 hours of Kyrgyz data -  0.003\% of the total pretraining data.
This shows that there is cross-lingual transfer that benefits low-resource languages and that additional capacity is important to realize this effect.

Chinese improves the least and gains are particularly large for languages for which the training corpus of \xlsrp{} contains more data due to VoxPopuli, e.g., for Swedish VP-400K adds more than 16K hours of unannontated speech and performance improves from 12.2 PER to 5.5 PER when comparing XLSR-53 to \xlsrpb{1}.
Finally, \xlsrpb{2} performs slightly better than \xlsrpb{1} on average with some languages improving while as others are performing slightly worse. 
The modest average improvement is likely because error rates are already low on this benchmark.

\insertCVtable

\subsubsection{Multilingual LibriSpeech}
\label{sec:res_mls}

Multilingual LibriSpeech is a common benchmark for evaluating multilingual speech recognition on eight European languages.
We consider a setup where we use ten hours of labeled data for each language~\citep{conneau2021xlsr} and compare to the prior work of~\citet{pratap2020mls} which uses all available labeled data as well as XLSR-53~\citep{conneau2021xlsr} which uses the same ten hour reduced labeled data setup. 

\autoref{tab:mls} shows that \xlsrp{} can outperform XLSR-53 on average by 1 WER at equal capacity and that additional model capacity results in an improvement of 2.9 WER on average for \xlsrpb{1}.
This result rivals the performance of the supervised models of~\citet{pratap2020mls} which is based on significantly more labeled data compared to the ten hour setup of \xlsrp{}.
Finally, on average \xlsrpb{2} does not show improvements over \xlsrpb{1}, which is similar to CommonVoice.

\insertMLStable

\subsubsection{VoxPopuli}
\label{sec:res_vp}

The VoxPopuli corpus provides about 1.8K hours of labeled speech data in 14 languages, ranging from 543 hours for English to 35 hours for Slovakian, as well as about 400K hours of unlabeled speech. 
This dataset is representative of a setting where a lot of unannotated data in the same domain as the labeled data is available.
We compare to the work of \citet{wang2021voxpopuli} which used a cross-lingually pretrained wav2vec 2.0 Base model on an earlier version of VoxPopuli that contained about 10K hours of unlabeled speech.

\autoref{tab:vp_asr} shows that cross-lingual pretraining (VP-10K) reduces WER from an average of 37.5 for supervised-only training (No pretraining) to 15.3 WER.
\xlsrp{} uses a lot more unlabeled VoxPopuli data and this results in improved performance: \xlsrpb{0.3} improves over VP-10K by an average of 2.5 WER.
The largest gains are on English, where WER improves from 16.2 to 10.2, likely due to the use of more English data from MLS during pretraining (over 44K hours).
Increasing model capacity to 1B parameters results in even better performance, reducing WER from an average of 15.3 for VP-10K to 10.6.

\insertVPtable

\subsubsection{LibriSpeech}
\label{sec:res_ls}

On LibriSpeech English ASR, we compare \xlsrpb{0.3} and \xlsrpb{1} to the wav2vec 2.0 English baseline. We see in Table~\ref{tab:ls} that with the same capacity and same fine-tuning procedure, the English wav2vec 2.0 significantly outperforms the \xlsrpb{0.3} in all data regimes, showing the capacity dilution and interference problem of our multilingual model. However, when increasing the capacity, the model is able to catch up with the monolingual results. In particular, \xlsrpb{1} outperforms wav2vec 2.0 LV-60k on the 10 minute setting, but is at a disadvantage in the 10 hour setting, where the English-focused monolingual model outperforms it by 0.7 WER on average. This suggests that higher-capacity models can circumvent the interference problem and can get strong results on high-resource languages, while still leveraging their cross-lingual transfer ability for lower-resource languages~\cite{conneau2021xlsr}.

\insertLStable


\subsection{Speech Classification}

Finally, we evaluate our approach on two speech utterance classification tasks, language identification and speaker identification. For these tasks we use our smallest model as these tasks require less capacity given the lower complexity of the tasks compared to the structured prediction problems of speech recognition and speech translation.

\subsubsection{Language Identification}

For language identification we adopt the setup of VoxLingua107~\citep{valk2020voxlingua107} which provides data for 107 different languages.
We train our model on the official train set, and report results on the development set, comprising 33 languages. 

\autoref{tab:lid_voxlingua} shows that our best model outperforms previous work, improving the best known prior work of~\citet{ravanelli2020pasep} by 1\% absolute, a relative error reduction of 15\%. 
For comparison, we also fine-tune the English-only wav2vec 2.0 pretrained on Libri-Light which performs surprisingly well on this multilingual task but is outperformed by the \xlsrp{} model by 1.5\% error rate on average.

\insertVLTable

\subsubsection{Speaker Identification}

Finally, we consider speaker identification on VoxCeleb1 where we fine-tune our model to distinguish between a fixed set of speakers given an utterance.
We compare to prior work including results published as part of the recently introduced SUPERB benchmark~\citep{yang2021superb} but note that their results are not strictly comparable because they do not fine-tune the underlying pre-trained model. 
All parameters of \xlsrp{} are fine-tuned, similar to the evaluation of all other tasks.
The results (\autoref{tab:voxceleb1}) show that our cross-lingual model also performs very well for speaker identification, even though utterances are mostly in English.

\insertVoxCelebOneTable

\subsection{Discussion}

\paragraph{Single model.}
Cross-lingual training results in a single model for multiple languages compared to a separate model for each language.
Training a cross-lingual model requires more effort than a single monolingual model but the resulting model can be used for many different languages.
Advances in architectures and training can also be deployed more easily since we only need to retrain a single model rather than many different ones.

In terms of accuracy, prior work in self-supervised learning for speech established that cross-lingually pretrained models are very competitive to monolingually pretrained models for speech recognition~\citep{conneau2021xlsr}.
Our experiments show a similar trend for speech-translation: \xlsrp{} can perform very competitively to English-only pretrained models for \englishx{} speech translation where the encoder only needs to encode English speech - a setting which favors monolingually pretrained models.

\paragraph{Performance trends.}
Overall, \xlsrp{} performs best for low-resource and mid-resource languages. 
For speech translation, we observe strong improvements for low- and mid-resource \xenglish{} directions and comparatively smaller gains on high-resource directions. 
Many low-resource directions which previously had performance in the 1-5 BLEU range improve to over 10-20 BLEU due to the better cross-lingual speech representations.
For \englishx{} directions, large enough cross-lingual models can even surpass the performance of English-only pretrained models.

Similarly, for speech recognition, we see strong improvements on BABEL, CommonVoice and VoxPopuli, benchmarks which include low- and mid-resource tasks.\footnote{MLS is a notable exception and we attribute the different performance pattern to prior work having pretrained on large amounts of in-domain data.}
We find that models trained on more data from more languages can perform as well or better than comparable models of the same size and we we observe this trend across all speech recognition benchmarks.
Keeping everything else equal, larger capacity models often further improve performance.

\section{Conclusion}

\xlsrp{} is a new self-supervised cross-lingual speech representation model which scales the number of languages, the amount of training data as well as model size.
The training corpus is an order of magnitude larger than prior work and covers 128 languages in 436K hours of recorded speech audio.
The resulting model enables state of the art results for \xenglish{} speech translation on CoVoST-2, outperforming prior art by a sizeable margin with the largest improvements on mid- and low-resource directions. 
It also performs competitively to the best \englishx{} work, without the use of equally applicable techniques such as self-training and language model decoding.

On speech recognition, \xlsrp{} sets a new state of the art on CommonVoice, VoxPopuli, several languages of BABEL, while performing competitively on MLS with much less labeled data.
These datasets cover a wide range of languages, data regimes and domains, demonstrating the generalization ability of \xlsrp{}. 
Our model also sets a new state of the art on the VoxLingua107 language identification benchmark. 
The largest \xlsrp{} model comprises 2B parameters which enables it to outperform a strong English-only pretrained model on \englishx{} speech translation, a setting which favors monolingually pretrained models. 
This shows that cross-lingually trained models with sufficient capacity can perform as well as specialized monolingually pretrained models.
We hope \xlsrp{} will help catalyze research in speech technology for many more languages of the world.
Models and code are publicly available on several platforms.

\subsubsection*{Acknowledgments}
We thank Morgane Riviere, Ann Lee, Anne Wu, Chaitanya Talnikar, Daniel Haziza, Mary Williamson, Emmanuel Dupoux for early access to VP-400K, and Christophe Ropers for his advise on language categorization. 
We also thank Min Xu, Jacob Kahn, Shruti Bhosale, Anjali Sridhar, and Tatiana Likhomanenko for help with infrastructure.
\bibliography{refs}
\bibliographystyle{iclr2022_conference}

\clearpage
\appendix
\section*{Appendix}
\section{Detailed Speech Translation Results}
\label{app:st_full}
\insertCovostXEnFull
\insertCovostEnXFull

\end{document}